\documentclass[letterpaper]{article} 
\usepackage{aaai2026}  
\usepackage{times}  
\usepackage{helvet}  
\usepackage{courier}  
\usepackage[hyphens]{url}  
\usepackage{graphicx} 
\usepackage{natbib}  
\usepackage{caption} 
\usepackage{algorithm}
\usepackage{algorithmic}

\usepackage{newfloat}
\usepackage{listings}
\DeclareCaptionStyle{ruled}{labelfont=normalfont,labelsep=colon,strut=off} 
\floatstyle{ruled}
\newfloat{listing}{tb}{lst}{}
\floatname{listing}{Listing}
\title{Beyond the Lower Bound: Bridging Regret Minimization and Best Arm Identification in Lexicographic Bandits}
\author {
    Bo Xue\textsuperscript{\rm 1,2},
    Yuanyu Wan\textsuperscript{\rm 3},
    Zhichao Lu\textsuperscript{\rm 1},
    Qingfu Zhang\textsuperscript{\rm 1,2}\thanks{Qingfu Zhang is the corresponding author.}
}
\affiliations {
    \textsuperscript{\rm 1} Department of Computer Science, City University of Hong Kong, Hong Kong, China \\
    \textsuperscript{\rm 2} The City University of Hong Kong Shenzhen Research Institute, Shenzhen, China\\
    \textsuperscript{\rm 3} School of Software Technology, Zhejiang University, Ningbo, China\\
    boxue4-c@my.cityu.edu.hk, wanyy@zju.edu.cn, \{zhichao.lu, qingfu.zhang\}@cityu.edu.hk
}

\usepackage{bibentry}

\usepackage{subfigure}
\usepackage{booktabs}

\usepackage{multirow}
\usepackage{bm}
\usepackage{float}
\usepackage{algorithm,algorithmic}

\usepackage{subfigure}
\usepackage{paralist,amssymb,mathtools}
\usepackage{multirow}
\usepackage{bm}
\usepackage{float}
\usepackage{color}
\usepackage{pifont}
\usepackage{makecell}
\usepackage{threeparttable}

\newtheorem{thm}{Theorem}

\newtheorem{lem}{Lemma}
\newtheorem{cor}{Corollary}

\newtheorem{defi}{Definition}

\def \E {\mathrm{E}}

\def \R {\mathbb{R}}
\def \N {\mathbb{N}}
\def \ID {\mathbb{I}}

\def \A {\mathcal{A}}
\def \S {\mathcal{S}}
\def \O {\mathcal{O}}
\def \event {\mathcal{E}}

\def \r{\bm{r}}

\DeclareMathOperator*{\argmax}{argmax}

\begin{document}

\maketitle

\begin{abstract}
In multi-objective decision-making with hierarchical preferences, lexicographic bandits provide a natural framework for optimizing multiple objectives in a prioritized order. In this setting, a learner repeatedly selects arms and observes reward vectors, aiming to maximize the reward for the highest-priority objective, then the next, and so on. While previous studies have primarily focused on regret minimization, this work bridges the gap between \textit{regret minimization} and \textit{best arm identification} under lexicographic preferences. We propose two elimination-based algorithms to address this joint objective. The first algorithm eliminates suboptimal arms sequentially, layer by layer, in accordance with the objective priorities, and achieves sample complexity and regret bounds comparable to those of the best single-objective algorithms. The second algorithm simultaneously leverages reward information from all objectives in each round, effectively exploiting cross-objective dependencies. Remarkably, it outperforms the known lower bound for the single-objective bandit problem, highlighting the benefit of cross-objective information sharing in the multi-objective setting. Empirical results further validate their superior performance over baselines.
\end{abstract}


\section{Introduction}
The multi-armed bandit (MAB) problem is a foundational framework for sequential decision-making under uncertainty \citep{Robbins:1952,Lai1985,Auer:2002}, with widespread applications in domains such as online recommendation systems \citep{Schwartz:2017}, clinical trials \citep{Villar:2015}, and adaptive routing \citep{Baruch:2008}. In the classical MAB setting \citep{Bandit:suvery}, a learner repeatedly selects one arm from a finite set of $K$ arms, each associated with an unknown reward distribution. Upon each selection, the learner observes a stochastic reward sampled from the distribution of the chosen arm. Depending on the learning objective, bandit algorithms are generally categorized into two primary paradigms: (1) regret minimization (RM), which aims to minimize the cumulative regret incurred by not always selecting the optimal arm \citep{Auer:2002:FAM,Abbasi:2011,Lykouris:2018}; and (2) best arm identification (BAI), which aims to identify the optimal arm using as few samples as possible \citep{Audibert:2010,Karnin:2013,Jamieson:2014,Emilie:2016,Jin:2024}.

While traditional bandit algorithms focus on optimizing a scalar reward \citep{Auer:2002:FAM}, many real-world applications involve multiple, often conflicting objectives \citep{Yutong:2021, Shu:2024}, which motivate the study of the multi-objective bandit problem \citep{Drugan:2013}. Several formulations have been proposed in this context, including scalarized regret minimization \citep{Saba:2015}, Pareto regret minimization \citep{Lu:2019-2, Xu:2023}, and Pareto set identification \citep{Auer:2016}. These methods offer different strategies for managing trade-offs among objectives, but generally assume that all objectives are equally important or can be aggregated into a single scalar value. However, in many practical scenarios, objectives have inherently different priorities. For instance, in medical diagnosis \citep{Faisal:2023}, patient safety typically outweighs considerations such as cost or treatment speed; in recommendation systems \citep{Li:2023}, fairness may be prioritized over user engagement.

An effective framework for modeling such hierarchical decision-making is lexicographic bandits \citep{Tekin:2018,Huyukt:2021}, where the agent seeks to optimize multiple objectives according to the lexicographic (i.e., priority-based) order. Unlike approaches that aggregate objectives into a single scalar using linear weights, the lexicographic bandit framework preserves the dominance structure: higher-priority objectives must be optimized before lower-priority ones are considered. This formulation provides a more faithful representation of structured decision-making in sensitive applications such as hyperparameter optimization \citep{Shaokun:2023} and multi-criteria resource allocation \citep{Kurokawa:2018}.

Research on lexicographic bandits has attracted increasing attention in recent years, with most studies focusing on the RM task \citep{Tekin:2018,Huyukt:2021,Xue:2024}. However, to the best of our knowledge, another significant task in the bandit literature, BAI, has not yet been explored in the context of lexicographic bandits. In many real-world scenarios, it is important to minimize regret during the learning phase while also accurately identifying the optimal arm at the end \citep{Zhong:2023}. For instance, in clinical trials, ethical considerations require providing effective treatments during the study (low regret), while the ultimate goal is to determine the most effective treatment (accurate BAI). These dual requirements motivate a central research question:

\begin{center}
\textit{Can we design algorithms for lexicographic bandits that effectively unify RM and BAI?}
\end{center}

In this work, we answer this question affirmatively and demonstrate that a unified treatment of RM and BAI in lexicographic bandits is not only possible, but also yields \textit{surprising benefits}. In particular, the rich multi-objective feedback naturally accelerates the elimination of suboptimal arms during the BAI process, thereby reducing the need to explore inferior actions and mitigating cumulative regret. This positive feedback loop between accurate identification and efficient learning highlights an unexpected advantage of jointly addressing BAI and RM in lexicographic bandits.

This paper presents the first algorithmic framework for lexicographic bandits that simultaneously tackles both RM and BAI tasks. Our main contributions are as follows:
\begin{itemize}

\item We propose a simple yet effective elimination-based algorithm, \textbf{LexElim-Out}, which sequentially filters suboptimal arms, starting from the highest-priority objective and proceeding to the lowest. This top-down elimination strategy ensures that lower-priority objectives are only considered after higher-priority objectives have been sufficiently optimized. Theoretically, LexElim-Out matches the best-known problem-dependent BAI guarantees for the primary objective, without compromising performance when optimizing additional objectives.

\item We further develop an enhanced algorithm, \textbf{LexElim-In}, which eliminates arms using joint reward information from all objectives in each round. By simultaneously incorporating information across objectives during each decision step, LexElim-In accelerates the identification and elimination of suboptimal arms. We show that it surpasses the known lower bounds for single-objective bandits in both regret and sample complexity, highlighting the advantage of exploiting the multi-objective structure.

\item LexElim-In also enjoys anytime performance guarantees. Specifically, we establish a minimax regret bound of $\widetilde{O}(\Lambda^i(\lambda)\cdot\sqrt{Kt})$ for each objective $i \in [m]$ at any round $t \geq 1$, ensuring that the regret grows at most at a square-root rate over time. This bound is comparable to the best-known results in single-objective bandits, while operating in a more challenging multi-objective setting.

\item Through extensive experiments on synthetic data, we demonstrate that both LexElim-Out and LexElim-In outperform existing baselines in cumulative regret and BAI sample complexity. Notably, LexElim-In exhibits superior performance on some instances, validating the benefit of joint exploitation of multi-objective reward signals.
\end{itemize}





\begin{table*}[t]
\begin{threeparttable}
\centering
\begin{tabular}{l l l c}
\toprule
 Algorithm & Sample Complexity & Regret Bound & \# Objectives \\
\midrule
\citet{Auer:2002:FAM} & -- & $\widetilde{O}\left(\sum_{\Delta(a)>0}\frac{1}{\Delta(a)}\right)$ & 1 \\
\citet{Degenne:2016}  & -- & $O\left(\sqrt{KT}\right)$ & 1 \\
\citet{Lattimore:2018} \textbf{(Lower Bound)}  & -- & $\Omega\left(\sum_{\Delta(a)>0}\frac{1}{\Delta(a)}\right)$ & 1\\

\citet{Karnin:2013} & $\widetilde{O}\left(\sum_{\Delta(a)>0}\frac{1}{(\Delta(a))^2}\right)$ & -- & 1\\
\citet{Jamieson:2014} \textbf{(Lower Bound)} & $\Omega\left(\sum_{\Delta(a)>0}\frac{1}{(\Delta(a))^2}\right)$ & -- & 1 \\
\citet{Degenne:2019} & $\widetilde{O}\left(\sum_{\Delta(a)>0}\frac{1}{(\Delta(a))^2}\right)$ & $\widetilde{O}\left(\sum_{\Delta(a)>0}\frac{1}{\Delta(a)}\right)$  & 1 \\
\midrule

LexElim-Out \textbf{(Ours)} & $\widetilde{O}\left(\sum_{j=1}^i\sum_{a\in \S(j)}\frac{1}{(\Delta^j(a))^2}\right)$ & $\widetilde{O}\left(\sum_{j=1}^i\sum_{a\in \S(j)}\frac{\Delta^i(a)}{(\Delta^j(a))^2}\right)$ & $i\in[m]$ \\
LexElim-In \textbf{(Ours)} & $\widetilde{O}\left(\sum_{\Delta^i(a)>0}\frac{1}{(\tilde{\Delta}(a))^2}\right)$ & \makecell[l]{$\widetilde{O}\left(\sum_{\Delta^i(a)>0}\frac{\Delta^i(a)}{(\tilde{\Delta}(a))^2}\right)$ \\ $\widetilde{O}\left(\Lambda^i(\lambda) \cdot \sqrt{KT}\right)$} & $i\in[m]$ \\
\bottomrule
\end{tabular}
\label{table}
\begin{tablenotes}
\item[1.] $\Delta^i(a) = \mu^i(a_*) - \mu^i(a)$ for all $a\in[K]$ and $i\in[m]$, where $a_*$ is the lex-optimal arm defined in Definition~2.
\item[2.] For single-objective works, we simplify the notation by letting $\Delta(a) := \Delta^1(a)$.
\item[3.] $\S(i)=\{a\in\O_*(i-1)\mid\Delta^i(a)>0\}, \O_*(i-1)=\{a\in[K]\mid\mu^j(a_*)=\mu^j(a),\ \forall j\in[i-1]\}$ and $\O_*(0)=[K]$.
\item[4.] $\tilde{\Delta}(a) = \max\limits_{i\in[m]} \left\{ \frac{\Delta^i(a)}{\Lambda^i(\lambda)} \cdot \mathbb{I}[\Delta^i(a) > 0] \right\}$, where $\Lambda^i(\lambda)=1+\lambda+\cdots+\lambda^{i-1}$ and $\lambda\geq0$ is defined in Eq.~\eqref{lex-identity-MAB}.
\end{tablenotes}
\end{threeparttable}
\caption{Overview of Our Results and Comparisons with RM and BAI Methods: Since $\tilde{\Delta}(a) \geq \Delta^1(a)$ for all $a \in [K]$, LexElim-In outperforms the lower bounds of the single-objective problem \citep{Jamieson:2014,Lattimore:2018}.}
\end{table*}

\section{Preliminaries}
This paper studies the lexicographic bandit problem, where a learner selects arms to simultaneously optimize multiple objectives that are ranked according to their importance.

Let $K\in\N_+$ denote the number of objectives, and $m\in\N_+$ be the number of objectives. For any $N \in \mathbb{N}_+$, let $[N]= \{1, 2, \ldots, N\}$ denote the index set. At each round $t \in[T]$, the learner chooses an arm $a_t \in [K]$ and receives a stochastic reward vector $\r_t(a_t) = [r_t^1(a_t), r_t^2(a_t), \ldots, r_t^m(a_t)] \in \R^m$. The component $r_t^i(a_t)$ corresponds to the reward for the $i$-th objective and is independently drawn from a $1$-sub-Gaussian distribution with an \textit{unknown} mean $\mu^i(a_t)\in[0,1]$. That is, for all $\beta \in \mathbb{R}$ and $i \in [m]$,
\begin{equation}\label{sub-Gaussian}
\E[e^{\beta r^i_t(a_t)}]\leq \text{exp}\left(\beta^2/2\right), \quad \mu^i(a_t)= \E[r_t^i(a_t)].
\end{equation}

The key challenge in lexicographic bandits is managing the hierarchical structure of objectives: the learner must optimize the most important objective first, followed by the second-most important, and so on. To formalize this, we adopt the standard notion of lexicographic dominance from prior work \citep{Huyukt:2021, Xue:2024}.

\begin{defi}[\textbf{Lexicographic Order}]
Let $a_1$, $a_2$ $\in [K]$ be two arms. We say that $a_1$ lexicographically dominates $a_2$ if there exists an index $i\in [m]$ such that $\mu^j(a_1)=\mu^j(a_2)$ for all $j < i$, and $\mu^i(a_1) > \mu^i(a_2)$.
\end{defi}

An illustrate example is that the arm with expected rewards $[5, 5, 2]$ lexicographically dominates the arm with expected rewards $[5, 4, 8]$, even though the latter has a higher value on the third objective. Lexicographic order induces a total order over arms, enabling the comparison of any two arms and thereby defining the notion of the lex-optimal arm.

\begin{defi}[\textbf{Lex-optimal Arm}]
An arm $a_*$ is lex-optimal if no other arm in $[K]$ lexicographically dominates it.
\end{defi}

We study two classical goals in the bandit literature, and adapt them to the lexicographic multi-objective setting. The first is \textbf{Regret Minimization (RM)}, which aims to minimize the cumulative regret for each objective over $T$ rounds,
\begin{equation*}
R^i(T) = T\cdot \mu^i(a_*) - \sum_{t=1}^T \mu^i(a_t), i \in [m].
\end{equation*}

The second is \textbf{Best Arm Identification (BAI)} with fixed confidence. Given a confidence level $\delta \in (0,1)$, the goal is to identify the optimal arm (or optimal arm set) with probability at least $1 - \delta$, using as few samples as possible.

Unlike the single-objective setting where the optimal arm is uniquely defined, in the multi-objective case, different objectives may induce different optimal arms. To capture this, we consider the following two types of optimal arm sets for each objective $i \in [m]$:
\begin{itemize}
\item $\O_*(i) = \{a \in [K] \mid \mu^j(a) = \mu^j(a_*) \text{ for all } j \in [i]\}$: the set of arms that match $a_*$ on the top $i$ objectives;
\item $\widetilde{\O}_*(i) = \{a \in [K] \mid \mu^i(a) \geq \mu^i(a_*)\}$: the set of arms that are optimal with respect to the $i$-th objective alone.
\end{itemize}


Let $T^i(\delta)$ and $\widetilde{T}^i(\delta)$ denote the number of samples used to identify $\O_*(i)$ and $\widetilde{\O}_*(i)$, respectively.  Thus, the sample complexity of identifying $a_*$ is $T^m(\delta)$ or $\max_{i \in [m]} \widetilde{T}^i(\delta)$. 

Finally, we introduce a parameter $\lambda$ to capture the trade-offs among conflicting objectives. In the lexicographic bandit problem, we assume that for any $i\geq2$ and $a\in[K]$,
\begin{equation}\label{lex-identity-MAB}
\mu^i(a)-\mu^i(a_*)\leq \lambda \cdot \max_{j\in[i-1]} \{\mu^j(a_*)-\mu^j(a)\}.
\end{equation}


\section{Related Work}

We review bandit work on four directions: regret minimization (RM), best arm identification (BAI), joint optimization of RM and BAI, and multi-objective bandits (MOB).


\subsubsection{RM.}
The seminal work of \citet{Robbins:1952} initiated the study of the MAB problem. A foundational algorithm for minimizing regret in stochastic MABs is the Upper Confidence Bound (UCB) algorithm \citep{Auer:2002:FAM}, which achieves a problem-dependent regret bound of $\widetilde{O}\left(\sum_{\Delta(a)>0} 1/\Delta(a)\right)$. To improve worst-case performance, \citet{Audibert:2009-MOSS} proposed the MOSS algorithm, which attains the minimax-optimal regret bound of $O(\sqrt{KT})$. This was further improved by \citet{Degenne:2016}, who developed an anytime variant of MOSS that removes the need for prior knowledge of the time horizon $T$, thereby improving its practicality. Additionally, \citet{Lattimore:2018} established a fundamental lower bound of $\Omega\left(\sum_{\Delta(a)>0} 1/\Delta(a)\right)$, highlighting the intrinsic complexity of the problem. These foundational results have been extended to structured bandit settings, such as linear bandits \citep{Linear:Bandit:08}, graphical bandits \citep{Alon:2015} and combinatorial bandits \citep{Chen:2016}.

\subsubsection{BAI.} 
Existing work on BAI can be categorized into two primary settings: (a) \textit{Fixed-confidence setting:} The algorithm aims to identify the best arm with probability at least $1-\delta$, using as few samples as possible. Early approaches include the Successive Elimination algorithm \citep{Eyal:2006}, which sequentially discards suboptimal arms based on empirical comparisons. Later works \citep{Karnin:2013,Garivier:2016} introduced more refined strategies that achieve near-optimal sample complexity by adaptively allocating samples to competitive arms. \citet{Jamieson:2014} established a lower bound showing that the sample complexity of any algorithm is at least $\Omega(\sum_{\Delta(a)>0}1/(\Delta(a))^2)$.

(b) \textit{Fixed-budget setting:} Given a fixed budget $T \in \N$, the objective is to minimize the probability of incorrect identification at time $T$. \citet{Audibert:2010} first studied this setting and designed an algorithm based on successive rejects, proved its optimality up to logarithmic factors. A subsequent work of \citet{Karnin:2013} further improved the theoretic guarantees, leaving only doubly-logarithmic gap. \citet{Carpentier:2016} constructed lower bounds to confirm the near-optimality of these results.


\subsubsection{RM and BAI.}
While RM and BAI have traditionally been treated as separate goals, recent studies have sought to address them jointly. \citet{Degenne:2019} explored both goals with a fixed confidence and introduced an algorithm $\text{UCB}_\alpha$, where the parameter $\alpha>1$ controls the trade-off between regret and sample complexity. Subsequently, \citet{Zhong:2023} quantified the trade-off between RM and BAI in the fixed-budget setting. In parallel, \citet{Qining:2023} developed algorithms that achieve asymptotic regret optimality in Gaussian bandit models. Most recently, \citet{Junwen:2024} established an information-theoretic lower bound for BAI with minimal regret and proposed an algorithm that attains asymptotic optimality.

\subsubsection{MOB.}
Multi-objective bandits aim to balance competing objectives, often without a unique optimal solution. Prior research has explored various notions of optimality and preference structures to address this challenge. Early studies focus extended the Pareto optimality concept to online learning \citep{Auer:2016,Kone:2024,Crepon:2024}, where the learner aims to approximate the Pareto front. Another line of work employs scalarization techniques \citep{Drugan:2013,Saba:2015,Wanigasekara:2019} to guide learning, based on utility functions or user-specified preferences. Lexicographic bandits, a specific form of preference-based MOB, have been studied under the RM framework \citep{Huyukt:2021,Tekin:2019,Xue:2024}. Our work contributes the first unified framework that simultaneously addresses RM and BAI under lexicographic preference, and we theoretically demonstrates how joint rewards signals lead to improved performance.


\section{Algorithms}
In this section, we propose two algorithms tailored for lexicographic bandits: LexElim-Out and LexElim-In. Both algorithms are based on the principle of arm elimination, but differ in how they utilize multi-objective information.

\subsection{Warm-up: LexElim-Out}

We begin by introducing LexElim-Out, a warm-up algorithm for the lexicographic MAB problem. This algorithm follows an outer-layer elimination strategy, where arms are pruned layer-by-layer according to the lexicographic priority of objectives. Details are provided in Algorithm~\ref{LexElim-Out}.


LexElim-Out requires prior knowledge of $|\O_*(i)|$, i.e., the number of arms that are optimal up to objective $i \in [m]$. This aligns with common practices in the single-objective BAI literature \citep{Bubeck:2009, Audibert:2010, Qining:2023}, where the optimal arm is typically assumed to be unique. Therefore, our setting does not require any additional information beyond what is standard in the single-objective BAI methods.

Given a confidence parameter $\delta \in (0,1)$, the number of arms $K$, the number of objectives $m$, and the cardinalities $|\O_*(i)|$ for all $i \in [m]$, LexElim-Out proceeds as follows. For each arm $a \in [K]$ and objective $i \in [m]$, it initializes the empirical mean reward $\hat{\mu}^i(a)$ and pull count $n(a)$ to zero, and the confidence width $c(a)$ to $+\infty$. The active arm set is initialized as $\A_1 = [K]$, and the round index as $t = 1$.

\begin{algorithm}[tb]
    \caption{Outer-layer Active Arm Elimination in Lexicographic Bandits (LexElim-Out)}
    \label{LexElim-Out}
    \begin{algorithmic}[1]
    \REQUIRE $\delta\in(0,1), K, m, \{|\O_*(i)|, \forall i\in[m]\}$
    \STATE Initialize empirical mean $\hat{\mu}^i(a)=0$, counter $n(a)=0$, and confidence width $c(a)=+\infty$ for $i\in[m]$, $a\in[K]$
    \STATE Initialize active set $\A_1=[K]$ and round counter $t=1$
    \FOR {$i=1,2,\ldots,m$}
        \WHILE {$|\A_t|>|\O_*(i)|$}
            \STATE Choose the arm $a_t=\argmax_{a\in\A_t}c(a)$
            \STATE $\hat{a}_t^i = \argmax_{a\in\A_t} \hat{\mu}^i(a)$
            \STATE $\A_{t+1} = \{a\in \A_t | \hat{\mu}^i(\hat{a}_t^i)-\hat{\mu}^i(a)\leq 2c(a_t)\}$
            \STATE Play $a_t$ and observe reward vectors $\r_t(a_t)$
            \STATE Update $\hat{\mu}^i(a_t)$ for all $i\in [m]$ by Eq.~\eqref{mean-reward}
            \STATE Update $n(a_t)$ and $c(a_t)$ by Eq.~\eqref{conf-term}
            \STATE Set $t=t+1$
        \ENDWHILE
    \ENDFOR
    \STATE \textbf{Output} the arm in $\A_{t}$
    \end{algorithmic}
\end{algorithm}

Then, LexElim-Out performs iterations over the objectives in order of priority, from the most to the least important. For each objective $i\in[m]$, it repeatedly performs elimination rounds until the size of the active arm set is reduced to the known optimal set size, i.e., $|\A_t|=|\O_*(i)|$. In each round, the algorithm selects the arm with the highest uncertainty,
\begin{equation*}
a_t = \argmax_{a \in \A_t} c(a).
\end{equation*}
It then identifies the empirical best arm with respect to the current objective, i.e., $\hat{a}_t^i = \arg\max_{a \in \A_t} \hat{\mu}^i(a)$. The active arm set is updated by retaining only those arms whose empirical means are within $2c(a_t)$ of the best empirical arm $\hat{a}_t^i$, 
\begin{equation*}
\A_{t+1} = \{ a \in \A_t \mid \hat{\mu}^i(\hat{a}_t^i) - \hat{\mu}^i(a) \leq 2c(a_t) \}.
\end{equation*}
This ensures that arms that are suboptimal on the $i$-th objective are eliminated.

After the elimination step, LexElim-Out plays the most uncertain arm $a_t$ and observes its reward vector $\r_t(a_t) = [r_t^1(a_t), r_t^2(a_t), \ldots, r_t^m(a_t)]$. The empirical mean for each objective $i \in [m]$ is updated using an incremental average,
\begin{equation}\label{mean-reward}
\hat{\mu}^i(a_t)=\frac{n(a_t)\cdot\hat{\mu}^i(a_t)+r_t^i(a_t)}{n(a_t)+1}.
\end{equation}
Next, the pull count $n(a_t)$ is incremented, and the confidence width $c(a_t)$ is updated by a concentration inequality,
\begin{equation}\label{conf-term}
\begin{aligned}
n(a_t)&=n(a_t)+1, \\
c(a_t)&=\sqrt{\frac{4}{n(a_t)}\log\left(\frac{6Km\cdot n(a_t)}{\delta}\right)}.
\end{aligned}
\end{equation}
The round index $t$ is then incremented to $t+1$. 

Once all objectives have been processed, LexElim-Out terminates and outputs the sole remaining arm in the final active set. The regret bounds and sample complexity of the algorithm are established in Theorems~\ref{thm:1} and~\ref{thm:2}, respectively. 

\begin{thm}\label{thm:1}
Suppose that Eq.~\eqref{sub-Gaussian} holds, define $\mathcal{S}(i) = \{a \in \mathcal{O}_*(i-1) \mid \Delta^i(a) > 0\}$ with $\mathcal{O}_*(0) = [K]$, and set $\gamma^i(\delta) = 64\log\left(\frac{392Km}{(\Delta^i(a))^2 \cdot \delta}\right)$. With probability at least $1-\delta$, for any objective $i\in[m]$, the regret of LexElim-Out satisfies
\begin{equation*}
R^i(t)\leq\sum_{j=1}^i\sum_{a\in \S(j)}\frac{\gamma^j(\delta)\cdot\Delta^i(a)}{(\Delta^j(a))^2}.
\end{equation*}
\end{thm}
\noindent\textbf{Remark~1} Theorem~\ref{thm:1} states that LexElim-Out achieves a regret bound of $\widetilde{O}\left(\sum_{j=1}^i \sum_{a \in \S(j)} \frac{\Delta^i(a)}{(\Delta^j(a))^2}\right)$ for any objective $i \in [m]$, with the following key implications.

\begin{itemize}
    \item For the primary objective ($i=1$), its regret bound is $\widetilde{O}\left(\sum_{\Delta^1(a)>0} \frac{1}{\Delta^1(a)}\right)$, matching the known lower bound for single-objective bandits \citep{Lattimore:2018}. This ensures no performance degradation for the highest-priority objective when optimizing additional objectives.
    
    \item For the secondary objective ($i=2$), its regret bound includes two terms:
    \begin{equation*}
        \underbrace{\widetilde{O}\left(\sum_{\Delta^1(a)>0} \frac{\Delta^2(a)}{(\Delta^1(a))^2}\right)}_{\text{cross-objective cost}} + \underbrace{\widetilde{O}\left(\sum_{a \in \S(2)} \frac{1}{\Delta^2(a)}\right)}_{\text{single-objective term}}.
    \end{equation*}
    The second term aligns with the regret bound in the single-objective setting. The first term captures the cost incurred on the second objective due to the need to prioritize the first objective. This cost becomes negligible if $(\Delta^1(a))^2 \gg \Delta^2(a)$, i.e., when arm $a$ is clearly suboptimal on the first objective and thus quickly eliminated.
\end{itemize}
The same decomposition can be applied to $i > 2$, where the regret bound includes cumulative cross-objective costs from all higher-priority objectives $j < i$, and a local term that matches the single-objective bound for objective $i$.



\begin{thm}\label{thm:2}
Suppose the same conditions and notations as in Theorem~\ref{thm:1}. With probability at least $1-\delta$, for any objective $i\in[m]$, the number of samples required by LexElim-Out to identify $\O_*(i)$ satisfies
\begin{equation*}
T^i(\delta)\leq \sum_{j=1}^i\sum_{a\in \S(j)}\frac{\gamma^j(\delta)}{(\Delta^j(a))^2}.
\end{equation*}
\end{thm}
\noindent\textbf{Remark~2} From Theorem~\ref{thm:2}, LexElim-Out identifies the optimal arm set for the first $i$ objectives using at most $\widetilde{O}\left(\sum_{j=1}^i \sum_{a \in \mathcal{S}(j)} \frac{1}{(\Delta^j(a))^2}\right)$ samples. In particular, for the highest-priority objective ($i = 1$), the sample complexity simplifies to $\widetilde{O}\left(\sum_{\Delta^1(a) > 0} \frac{1}{(\Delta^1(a))^2}\right)$, which matches the known lower bound for single-objective bandits \citep{Jamieson:2014}. This implies that LexElim-Out identifies the optimal arm for the primary objective as efficiently as state-of-the-art single-objective algorithms \citep{Karnin:2013}. For general $i \in [m]$, the bound reflects that identifying the lex-optimal arm requires solving a sequence of BAI problems, where suboptimal arms for higher-priority objectives are progressively eliminated before being evaluated on lower-priority ones.

\subsection{Improved Algorithm: LexElim-In}



LexElim-Out handles objectives layer by layer, it ignores lower-priority objectives when optimizing higher-priority ones. As a result, the arm selection for lower-priority objectives in early rounds is purely random, lacking any targeted exploration. To address this limitation, we propose an improved algorithm, LexElim-In, which adopts an inner-layer elimination strategy that leverages information from all objectives throughout the decision-making process. The complete procedure is presented in Algorithm~\ref{LexElim-In}.


Given a confidence level $\delta \in (0,1)$, the number of arms $K$, the number of objectives $m$, and a trade-off parameter $\lambda \geq 0$, LexElim-In begins with an initialization phase similar to that of LexElim-Out. Specifically, for each arm $a \in [K]$ and each objective $i \in [m]$, the empirical mean reward $\hat{\mu}^i(a)$ and pull count $n(a)$ are set to zero, and the confidence width $c(a)$ is initialized to $+\infty$. The initial active set of arms is defined as $\A_1 = [K]$, and the round index is initialized as $t = 1$.

\begin{algorithm}[tb]
    \caption{Inner-layer Active Arm Elimination in Lexicographic Bandits (LexElim-In)}
    \label{LexElim-In}
    \begin{algorithmic}[1]
    \REQUIRE $\delta\in(0,1), K, m, \lambda\geq0$
    \STATE Initialize empirical mean $\hat{\mu}^i(a)=0$, counter $n(a)=0$, and confidence width $c(a)=+\infty$ for $i\in[m]$, $a\in[K]$
    \STATE Initialize active set $\A_1=[K]$ and round counter $t=1$
    \WHILE {$|\A_t|>1$}
        \STATE Choose the arm $a_t=\argmax_{a\in\A_t}c(a)$
        \STATE Initialize the arm set $\A_t^0 = \A_t$
        \FOR {$i=1,2,\ldots,m$}
            \STATE $\hat{a}_t^i = \argmax_{a\in\A_t^{i-1}} \hat{\mu}^i(a)$
            \STATE $\A_t^i = \{a\in \A_t^{i-1} | \hat{\mu}^i(\hat{a}_t^i)-\hat{\mu}^i(a)\leq(2+4\lambda+\cdots+4\lambda^{i-1})\cdot c(a_t)\}$
        \ENDFOR
        \STATE Play $a_t$ and observe reward vectors $\r_t(a_t)$
        \STATE Update $\hat{\mu}^i(a_t)$ for all $i\in [m]$ by Eq.~\eqref{mean-reward}
        \STATE Update $n(a_t)$ and $c(a_t)$ by Eq.~\eqref{conf-term}
        \STATE Update $\A_{t+1}=\A_{t}^m$ and $t=t+1$
    \ENDWHILE
    \STATE \textbf{Output} the arm in $\A_{t}$
    \end{algorithmic}
\end{algorithm}

At each round, LexElim-In selects the arm $a_t \in \A_t$ with the largest confidence width $c(a)$, corresponding to the highest uncertainty, and plays this arm. It then updates the active arm set through a layered filtering process that incorporates empirical means across all objectives in a nested fashion.

Specifically, let $\A_t^0 = \A_t$ and for each objective $i = 1, 2, \ldots, m$, LexElim-In identifies the empirical best arm $\hat{a}_t^i = \arg\max_{a \in \A_t^{i-1}} \hat{\mu}^i(a)$, and eliminates arms in $\A_t^{i-1}$ whose empirical mean falls below that of $\hat{a}_t^i$ by more than a scaled confidence threshold. Formally, the updated set is
\begin{equation}\label{elimination-step}
\begin{aligned}
\A_t^i &= \left\{ a \in \A_t^{i-1} \mid \hat{\mu}^i(\hat{a}_t^i) - \hat{\mu}^i(a) \leq \right.\\
&\quad\left.\left(2 + 4\lambda + \cdots + 4\lambda^{i-1} \right) \cdot c(a_t) \right\}.
\end{aligned}
\end{equation}
The scaling factor $2 + 4\lambda + \cdots + 4\lambda^{i-1}$ grows geometrically with $i$, allowing lower-priority objectives to tolerate larger reward gaps while still contributing to elimination decisions.

After completing the elimination process across all $m$ objectives, LexElim-In updates the active set to $\A_{t+1} = \A_t^m$. It then pulls arm $a_t$ to observe the full reward vector $\r_t(a_t) = [r_t^1(a_t), \ldots, r_t^m(a_t)]$. For each objective $i \in [m]$, the empirical mean $\hat{\mu}^i(a_t)$ is updated using an incremental average defined in Eq.~\eqref{mean-reward}. The pull count $n(a_t)$ and the confidence width $c(a_t)$ are then updated by Eq.~\eqref{conf-term}. The round index is incremented, and the procedure repeats until the active set contains only a single arm.



The key innovation of LexElim-In is its \textit{cross-objective elimination} strategy, which utilizes information from all objectives at each round. By jointly incorporating elimination evidence across objectives, LexElim-In more efficiently eliminates suboptimal arms, especially when lower-priority objectives provide stronger signals. This approach leads to faster identification of the lexicographic optimum compared to LexElim-Out, albeit at the cost of requiring the prior knowledge $\lambda$. Formal regret and sample complexity guarantees are presented in Theorems~\ref{thm:3} and~\ref{thm:4}, respectively.

\begin{thm}\label{thm:3}
Suppose that Eq.~\eqref{sub-Gaussian} and Eq.~\eqref{lex-identity-MAB} hold. Define
$$\Lambda^i(\lambda)=\sum_{j=0}^{i-1}\lambda^j, \text{ and } \gamma^i(\delta)=64\log\left(\frac{392Km}{(\Delta^i(a))^2\cdot\delta}\right).$$
With probability at least $1-\delta$, for any objective $i\in[m]$, the regret of LexElim-In satisfies
\begin{equation*}
R^i(t)\leq \sum_{\Delta^i(a)>0}\min_{j\in[m]}\left\{\frac{(\Lambda^j(\lambda))^2\cdot \Delta^i(a)\cdot \gamma^j(\delta)}{(\Delta^j(a))^2\cdot\ID(\Delta^j(a)>0)}\right\}.
\end{equation*}
\end{thm}

\noindent\textbf{Remark~3} For the primary objective ($i=1$), the regret incurred due to $\Delta^1(a)>0$ is bounded by $$\min_{j\in[m]}\left\{\frac{\Delta^1(a)\cdot (\Lambda^j(\lambda))^2}{(\Delta^j(a))^2\cdot\ID(\Delta^j(a)>0)}\right\}\leq \frac{1}{\Delta^1(a)},$$ where the right-hand side matches the known lower bound \citep{Lattimore:2018}. The existence of $\min_{j\in[m]}$ allows the bound to go beyond the lower bound: if for some $j \geq 2$, the suboptimality gap $\Delta^j(a)$ is much larger than $\Delta^1(a) \cdot \Lambda^j(\lambda)$, the corresponding regret term can become significantly smaller than $1/\Delta^1(a)$. Thus, LexElim-In can adaptively exploit auxiliary objectives to accelerate learning.

Moreover, while the gap-dependent bound in Theorem~\ref{thm:3} highlights how LexElim-In can exploit the relative gap structures among objectives to reduce regret, it remains essential to understand the algorithm's behavior in the worst case.

\begin{cor}\label{cor:1}
Suppose the same conditions and notations as in Theorem~\ref{thm:3}. With probability at least $1-\delta$, for any objective $i\in[m]$, the regret of LexElim-In satisfies
\begin{equation*}
R^i(t)\leq \widetilde{O}\left(\Lambda^i(\lambda)\cdot \sqrt{Kt}\right).
\end{equation*}
\end{cor}
Corollary~\ref{cor:1} shows that for any objective $i\in[m]$, the worst-case regret of LexElim-In grows at most as $\widetilde{O}(\Lambda^i(\lambda) \sqrt{Kt})$. This matches the minimax bound $\widetilde{O}(\sqrt{Kt})$ of single-objective bandits~\citep{Degenne:2016}, up to the factor $\Lambda^i(\lambda)$. Hence, LexElim-In achieves minimax-optimal regret rates in terms of $K$ and $t$. Importantly, since $\Lambda^1(\lambda) = 1$, the regret for the highest-priority objective remains unaffected by the inclusion of lower-priority objectives, ensuring no performance degradation when optimizing multiple objectives simultaneously.

\begin{thm}\label{thm:4}
Suppose the same conditions and notations as in Theorem~\ref{thm:3}. With probability at least $1-\delta$, for any objective $i\in[m]$, the number of samples required by LexElim-In to identify $\widetilde{\O}_*(i)$ satisfies
\begin{equation*}
\widetilde{T}^i(\delta)\leq \sum_{\Delta^i(a)>0}\min_{j\in[m]}\left\{\frac{(\Lambda^j(\lambda))^2\cdot \gamma^j(\delta)}{(\Delta^j(a))^2\cdot\ID(\Delta^j(a)>0)}\right\}.
\end{equation*}
\end{thm}

\noindent\textbf{Remark~4} Theorem~\ref{thm:4} characterizes the sample complexity of LexElim-In for identifying the optimal arm set $\widetilde{\O}_*(i)$ for the $i$-th objective, revealing an objective-adaptive complexity. For each suboptimal arm $a$, the cost of distinguishing it is governed by the most distinguishable objective $j \in [m]$. In particular, if some objective $j$ exhibits a large suboptimality gap $\Delta^j(a)$ for a given arm $a$, that arm can often be eliminated early, without requiring extensive exploration of other objectives. In such case, LexElim-In adaptively leverages the reward structure across objectives to accelerate the identification process. Notably, in the single-objective setting, the lower bound on sample complexity is known to be $\Omega(\sum_{\Delta(a)>0} \frac{1}{(\Delta(a))^2})$~\citep{Jamieson:2014}. Our bound recovers this result when $i = 1$, since $\Lambda^1(\lambda) = 1$, and the $\min_{j \in [m]}$ term ensures our result surpasses this lower bound.

\subsubsection{Cross-objective Acceleration.}
\begin{figure}[tb]
\setlength{\abovecaptionskip}{0.1cm}
\setlength{\belowcaptionskip}{0cm}
\centering 
\includegraphics[width=0.45\textwidth]{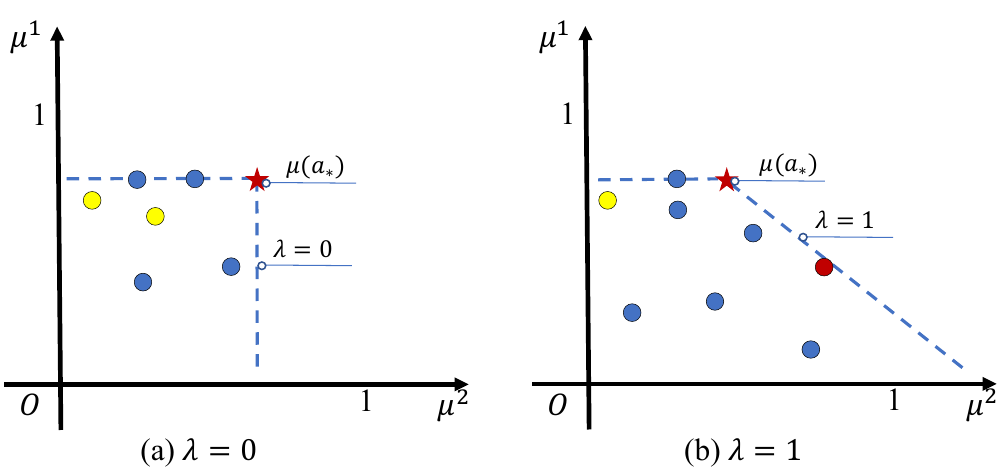}
\caption{Cross-objective Acceleration}
\label{fig:cross-infor}
\end{figure}

Figure~\ref{fig:cross-infor} illustrates how the second objective can accelerate BAI under varying degrees of trade-offs. The red star denotes the lex-optimal arm, while the circles represent suboptimal arms. In Figure~\ref{fig:cross-infor}(a), there is no conflict between other arms and the lex-optimal arm, resulting in $\lambda = 0$. The \textit{two yellow arms} exhibit much larger reward gaps in the second objective than in the first, enabling LexElim-In to efficiently eliminate them by leveraging second objective information. Figure~\ref{fig:cross-infor}(b) shows a conflict between the lex-optimal arm and the red suboptimal arm, leading to $\lambda = 1$. In this case, only \textit{the yellow arm that is far from the optimal arm} can be quickly eliminated, as the confidence term for the second objective is scaled by $2 + 4\lambda = 6$, as specified in Eq.~\eqref{elimination-step}.

\begin{figure*}[tb]
  \centering 
  \setlength{\abovecaptionskip}{0cm}
  \setlength{\belowcaptionskip}{0cm}
  \includegraphics[width=1\textwidth]{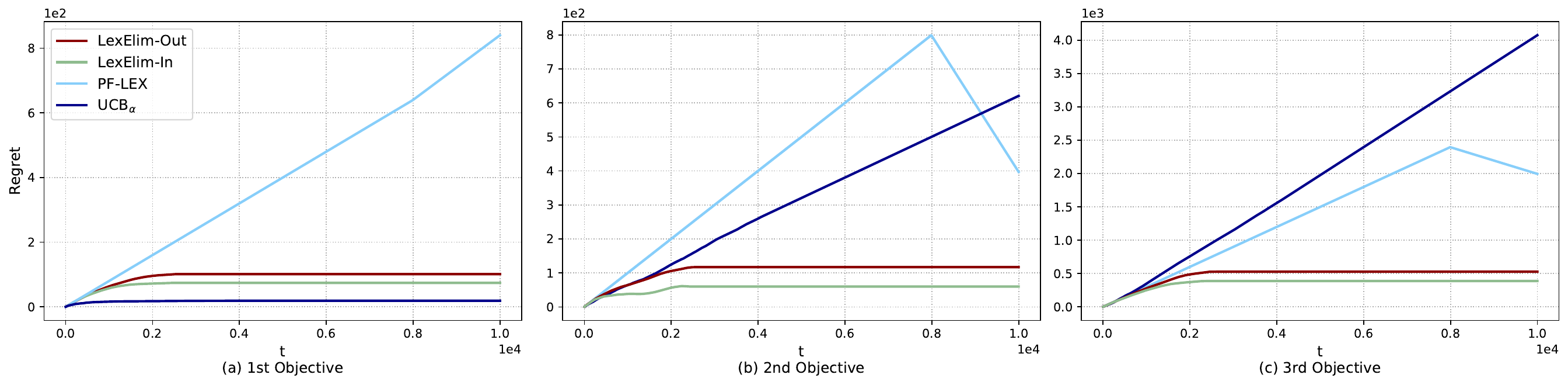}
  \caption{Regret Comparison of Our Algorithms versus PF-LEX and $\text{UCB}_\alpha$: $K=10$}
\label{fig:RM}
\end{figure*}

\begin{figure*}[tb]
  \centering 
  \setlength{\abovecaptionskip}{0cm}
  \setlength{\belowcaptionskip}{0cm}
  \includegraphics[width=1\textwidth]{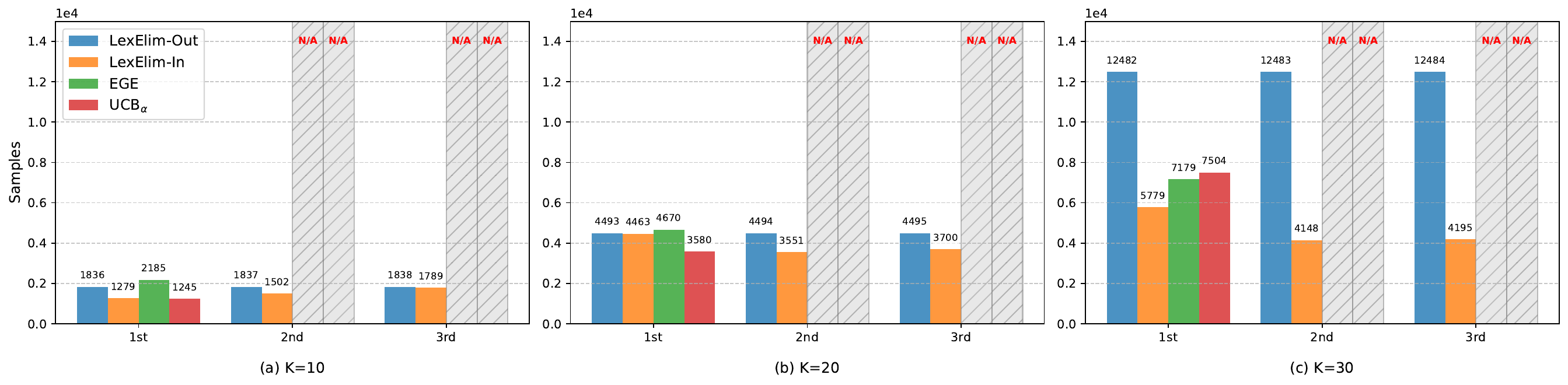}
  \caption{Sample Complexity Comparison of Our Algorithms versus EGE and $\text{UCB}_\alpha$}
\label{fig:BAI}
\end{figure*}

\section{Experiments}
In this section, we evaluate the empirical performance of our proposed algorithms, LexElim-Out and LexElim-In, on both RM and BAI tasks in lexicographic multi-objective bandits. Experiments are conducted on a Windows 10 laptop with Intel(R) Core(TM) i7-1170 CPU and 32GB memory.

\noindent\textbf{Baselines.} There are three baselines. The first is EGE, which addresses BAI in single-objective MAB \citep{Karnin:2013}. The second is $\text{UCB}_\alpha$, designed to handle both BAI and RM in the single-objective MAB setting \citep{Degenne:2019}. The third is PF-LEX, an algorithm tailored to lexicographic MAB, which focuses on the RM task \citep{Huyukt:2021}.

\noindent\textbf{Experimental Setup.} We consider settings with $m=3$. The expected rewards across the three objectives are defined as: $\mu^1(a)=1-\min_{p\in\{0.3,0.6,0.9\}}|a/K-p|$, $\mu^2(a)=1-2\times\min_{p\in\{0.5,0.8\}}|a/K-p|$, $\mu^3(a)=1-2\times|a/K-0.5|$, $a \in [K]$. This construction ensures that multiple arms are optimal for the higher-priority objectives: $\{0.3K, 0.6K, 0.9K\}$ are optimal for the first objective, while $\{0.6K, 0.9K\}$ are optimal for both the first and second objectives. To identify the unique lex-optimal arm $a_*=0.6K$, all three objectives must be considered. Stochastic rewards $r_t^i(a)$ are drawn from Gaussian distributions with mean $\mu^i(a)$ and variance $0.1$. Each algorithm is run for 10 independent trials, and we report the average regret and sample complexity.

\noindent\textbf{RM Results.} For those RM algorithms ($\text{UCB}_\alpha$, PF-LEX, LexElim-Out, and LexElim-In), we fix $K=10$ and run each algorithm for $T=10{,}000$. Figure~\ref{fig:RM} presents the cumulative regret over time, where Panels (a), (b), and (c) correspond to objectives 1, 2, and 3, respectively. LexElim-Out and LexElim-In exhibit uniformly sublinear regret growth across all objectives, demonstrating their ability to optimize multiple objectives simultaneously. In contrast, $\text{UCB}_\alpha$ tailored for single-objective optimization, only achieves low regret for the first objective, while incurring linear regret on the second and third. Although PF-LEX is designed for multi-objective settings, it lacks theoretical guarantees under general regret metrics and suffers from a slower convergence rate, as reflected in its $\widetilde{O}(T^{2/3})$ regret bound.

\noindent\textbf{BAI Results.} For BAI algorithms (EGE, $\text{UCB}_\alpha$, LexElim-Out, and LexElim-In), we set the confidence level $\delta = 0.01$ and evaluate their performance under varying numbers of arms $K \in \{10, 20, 30\}$. The results are shown in Figure~\ref{fig:BAI}, where Panels (a) – (c) correspond to increasing $K$. All algorithms require more samples as $K$ increases, reflecting the greater difficulty of distinguishing between arms when reward gaps shrink. LexElim-In consistently outperforms the baselines, and its advantage becomes more significant with larger $K$. This is because LexElim-In exploits information from lower-priority objectives, which have larger reward gaps and provide stronger signals for elimination. In our setting, the reward gaps for the second and third objectives are twice as large as that of the first, allowing LexElim-In to identify the optimal arm more efficiently.

\section{Conclusion and Future work}
This paper develops the first unified framework for simultaneously addressing both RM and BAI tasks in lexicographic multi-objective bandits. We propose two principled algorithms, LexElim-Out and LexElim-In, which adhere to the lexicographic preference structure while optimizing multiple objectives. LexElim-Out adopts a conservative elimination strategy that sequentially filters arms based on priority, ensuring no compromise on  higher-priority objectives. LexElim-In exploits the joint reward signals across all objectives to perform more efficient arm elimination. We provide a comprehensive theoretical analysis for both algorithms: LexElim-Out matches the known instance-dependent lower bounds for the primary objective, while LexElim-In achieves better instance-dependent bounds than classical single-objective methods.

An interesting direction for future work is to establish tighter lower bounds for lexicographic RM and BAI that explicitly capture the interactions among objectives. Additionally, eliminating the need for prior knowledge of the parameter $\lambda$ would further enhance the applicability of LexElim-In.

\bibliography{aaai2026}

\newpage

\onecolumn

\appendix

\section{Proof of Theorem~\ref{thm:1}}

For clarity, throughout the proof in appendix, we use the notations $\hat{\mu}^i_t(a)$, $n_t(a)$, and $c_t(a)$ to denote the values of $\hat{\mu}^i(a)$, $n(a)$, and $c(a)$ at the beginning of round $t$, respectively.

We first present a high-probability confidence interval for the expected rewards of all objectives.
\begin{lem}\label{lem:confidence-interval}
With probability at least $1-\delta$, for any $t\geq1$, 
\begin{equation*}
\left|\hat{\mu}_t^i(a)-\mu^i(a)\right|\leq c_t(a)=\sqrt{\frac{4}{n(a_t)}\log\left(\frac{6Km\cdot n(a_t)}{\delta}\right)}, i\in [m], a \in [K].
\end{equation*}
\end{lem}

This lemma provides a standard concentration inequality, bounding the deviation between the empirical and true rewards, and serves as a foundation for the subsequent analysis.

Let $\event$ denote the following high-probability event:
\begin{equation}\label{event}
\event = \left\{ \forall t \in [T],\ \forall a \in [K],\ \forall i \in [m]:\ \left| \hat{\mu}_t^i(a) - \mu^i(a) \right| \leq c_t(a) \right\}.
\end{equation}
By the argument in Lemma~\ref{lem:confidence-interval}, the event $\event$ holds with probability at least $1 - \delta$.

Next, we present three technical lemmas that characterize how many times an arm can be pulled before it is eliminated. These lemmas serve as analytical tools to facilitate the regret analysis.

The first lemma provides a useful inequality for comparing logarithmic expressions, which will be instrumental in simplifying bounds on the number of arm pulls.

\begin{lem}\label{lem:tool1}
Let $a > 0$, $b > 0$ and $ab>e$. If $x > a \log(ab)$, then $x > a \log(bx)$.
\end{lem}

The second lemma analyzes the behavior of the confidence radius function and shows that it decreases as the number of pulls increases.

\begin{lem}\label{lem:tool2}
Let $f(n) = 4\sqrt{ \frac{4}{n} \log\left( \frac{6Km \cdot n}{\delta} \right) }$ for $n > 0$. Then, $f(n)$ is strictly decreasing for all
\begin{equation*}
n > \frac{e\delta}{6Km}.
\end{equation*}
In particular, since $\frac{e\delta}{6Km} \ll 1$ in typical applications, the function $f(n)$ is strictly decreasing for all $n \geq 1$.
\end{lem}

The detailed proofs of these three technical lemmas are deferred to the end of this appendix. The following lemma shows that the number of times any two active arms have been pulled remains nearly balanced throughout the execution of Algorithm~\ref{LexElim-Out}.

\begin{lem}\label{lem:tool3}
In Algorithm~\ref{LexElim-Out}, for any arm $a_1,a_2\in\A_t$, their pull counts satisfy:
\begin{equation*}
n_t(a_1)-1\leq n_t(a_2)\leq n_t(a_1)+1.
\end{equation*}
\end{lem}

\noindent\textbf{Proof.} We prove this lemma by induction. At the initialization step, all arms have $n_1(a) = 0$, and the condition holds trivially.

Suppose at round $t$, for all $a_1, a_2 \in \A_t$, the pull counts satisfy $|n_t(a_1) - n_t(a_2)| \leq 1$. Now consider how the pull counts change at round $t$. LexElim-Out selects the arm $a_t$ as:
\begin{equation*}
a_t = \arg\max_{a \in \A_t} c(a),
\end{equation*}
where $c(a) = \sqrt{ \frac{4}{n(a)} \log\left( \frac{6Km \cdot n(a)}{\delta} \right) }$ is a strictly decreasing function of $n(a)$ (cf. Lemma~\ref{lem:tool2}).

Thus, at each round, LexElim-Out chooses the arm with the \textit{fewest number of pulls}. Let $n_{\min} := \min_{a \in \A_t} n_t(a)$. At round $t$, LexElim-Out chooses some arm $a_t$ with $n_t(a_t) = n_{\min}$ and increments its count:
\begin{equation*}
n_{t+1}(a_t) = n_t(a_t) + 1, \quad \text{while for all } a \neq a_t,\ n_{t+1}(a) = n_t(a).
\end{equation*}
After this round, the former minimum becomes $n_{\min} + 1$. All arms now have pull counts either $n_{\min}$ or $n_{\min} + 1$. Therefore, all arms' pull counts differ by at most $1$. This completes the induction.
$\hfill\square$

Equipped with the previous lemmas, we can now bound the number of times a suboptimal arm (with respect to the $i$-th objective) can be pulled before elimination.

\begin{lem}\label{lem:bounded-times}
Suppose the event $\event$ in \eqref{event} holds. For each objective $i \in [m]$, define:
\begin{itemize}
    \item $\O_*(i-1) := \left\{ a \in [K] \;\middle|\; \mu^j(a) = \mu^j(a_*)\ \text{for all } j \in [i-1] \right\}$;
    \item $\S(i) := \left\{ a \in \O_*(i-1) \;\middle|\; \Delta^i(a) > 0 \right\}$.
\end{itemize}
In Algorithm~\ref{LexElim-Out}, for any arm $a \in \S(i)$, the number of times it is played is at most
\begin{equation*}
n_t(a) \leq \frac{64}{(\Delta^i(a))^2} \log\left( \frac{392Km}{(\Delta^i(a))^2 \cdot \delta} \right).
\end{equation*}
\end{lem}

\noindent\textbf{Proof.} Let $a \in \S(i)$ and suppose it is eliminated at round $t$. By Algorithm~\ref{LexElim-Out}, the elimination condition is:
\begin{equation*}
\hat{\mu}^i_t(\hat{a}_t^i) - \hat{\mu}^i_t(a) > 2c_t(a_t) \quad \Longleftrightarrow \quad \hat{\mu}^i_t(\hat{a}_t^i)-c_t(a_t)> \hat{\mu}^i_t(a)+c_t(a_t),
\end{equation*}
where $a_t = \arg\max_{a \in \A_t} c_t(a)$ denotes the arm with the largest confidence width.

Since $\hat{a}_t^i=\argmax_{a\in\A_t}\hat{\mu}^i_t(\hat{a}_t^i)$ and $a_*\in \A_t$, a sufficient condition to eliminate $a$ is:
\begin{equation*}
\hat{\mu}^i_t(a_*)-c_t(a_t)> \hat{\mu}^i_t(a)+c_t(a_t).
\end{equation*}
Using the confidence event in \eqref{event}, we have that for all $t \geq 1$,
\begin{equation*}
|\hat{\mu}^i_t(a) - \mu^i(a)| \leq c_t(a)\leq c_t(a_t), \quad \text{for all } i \in [m],\ a \in [K].
\end{equation*}
Thus, it follows another sufficient condition to eliminate $a$:
\begin{equation*}
\mu^i(a_*) - 2c_t(a_t) > \mu^i(a) + 2c_t(a_t) \Leftrightarrow \Delta^i(a) = \mu^i(a_*) - \mu^i(a) > 4c_t(a_t).
\end{equation*}
Now recall that the confidence width is defined as
\begin{equation*}
c_t(a_t) = \sqrt{ \frac{4}{n_t(a_t)} \log\left( \frac{6Km \cdot n_t(a_t)}{\delta} \right) }.
\end{equation*}
Substituting into the inequality above, we obtain:
\begin{equation*}
\Delta^i(a) > 4 \sqrt{ \frac{4}{n_t(a_t)} \log\left( \frac{6Km \cdot n_t(a_t)}{\delta} \right) }.
\end{equation*}

By Lemma~\ref{lem:tool3}, the number of pulls among arms in $\A_t$ differs by at most one, i.e., $n_t(a_t)\geq n_t(a)-1$. Meanwhile, Lemma~\ref{lem:tool3} tells that $c_t(\cdot)$ is decreasing with respect to $n_t(a_t)$. Thus, a sufficient condition for eliminating $a$ becomes:
\begin{equation*}
\Delta^i(a) > 4 \sqrt{ \frac{4}{n_t(a)-1} \log\left( \frac{6Km \cdot (n_t(a)-1)}{\delta} \right) }.
\end{equation*}
Squaring both sides gives:
\begin{equation*}
(\Delta^i(a))^2 > \frac{64}{n_t(a)-1} \log\left( \frac{6Km \cdot (n_t(a)-1)}{\delta} \right).
\end{equation*}
Rewriting this inequality yields:
\begin{equation}\label{eq:lem4-2}
n_t(a)-1> \frac{64}{(\Delta^i(a))^2} \log\left( \frac{6Km \cdot (n_t(a)-1)}{\delta} \right).
\end{equation}
To obtain an explicit upper bound on $n_t(a)$, we apply Lemma~\ref{lem:tool1} with:
\begin{equation*}
a = \frac{64}{(\Delta^i(a))^2}, \quad b = \frac{6Km}{\delta}, \quad x = n_t(a).
\end{equation*}
According to Lemma~\ref{lem:tool1}, if
\begin{equation*}
n_t(a) > \frac{64}{(\Delta^i(a))^2} \log\left( \frac{384Km}{(\Delta^i(a))^2 \cdot \delta} \right) + 1,
\end{equation*}
then inequality~\eqref{eq:lem4-2} holds, which implies that arm $a$ will be eliminated at that point.

Therefore, the number of times arm $a$ is pulled is at most
\begin{equation*}
n_t(a) \leq \frac{64}{(\Delta^i(a))^2} \log\left( \frac{392Km}{(\Delta^i(a))^2 \cdot \delta} \right),
\end{equation*}
which completes the proof of Lemma~\ref{lem:bounded-times}.
$\hfill\square$

We now complete the proof of Theorem~\ref{thm:1}. Recall that the regret for each objective arises only from suboptimal arms that are not eliminated early enough. For the first objective, only the arms in $\S(1)$ incur regret, and by Lemma~\ref{lem:bounded-times}, each such arm is played at most
\begin{equation*}
\frac{64}{(\Delta^1(a))^2} \log\left( \frac{392Km}{(\Delta^1(a))^2 \cdot \delta} \right)=\frac{\gamma^1(\delta)}{(\Delta^1(a))^2}
\end{equation*}
times. Therefore, the total regret for the first objective is bounded by:
\begin{equation*}
R^1(t)\leq\sum_{a\in \S(1)}\frac{\gamma^1(\delta)}{\Delta^1(a)}.
\end{equation*}
For the second objective, regret may arise from both $\S(1)$ and $\S(2)$. Any arm $a \in \S(1)$ may continue to be pulled before being eliminated, thereby contributing regret proportional to $\Delta^2(a)$. Its regret contribution is bounded by:
\begin{equation*}
\frac{\gamma^1(\delta)\cdot\Delta^2(a)}{(\Delta^1(a))^2}.
\end{equation*}
Meanwhile, for arms $a \in \S(2)$, each is played at most
\begin{equation*}
\frac{64}{(\Delta^2(a))^2} \log\left( \frac{392Km}{(\Delta^2(a))^2 \cdot \delta} \right)=\frac{\gamma^2(\delta)}{(\Delta^2(a))^2}
\end{equation*}
times, incurring regret at most $\frac{\gamma^2(\delta)}{\Delta^2(a)}$ each. Hence, the total regret for the second objective satisfies:
\begin{equation*}
R^2(t)\leq\sum_{a\in \S(1)}\frac{\gamma^1(\delta)\cdot\Delta^2(a)}{(\Delta^1(a))^2}+\sum_{a\in \S(2)}\frac{\gamma^2(\delta)}{\Delta^2(a)}.
\end{equation*}

By the same reasoning, for the $i$-th objective ($i \in [m]$), regret may be contributed by all arms in $\S(1), \dots, \S(i)$. Specifically, an arm $a \in \S(j)$ contributes regret to the $i$-th objective as long as it is not eliminated before stage $j$, and is pulled while optimizing objectives $1$ through $j$. Each such arm contributes at most
\begin{equation*}
\frac{64\cdot\Delta^i(a)}{(\Delta^j(a))^2} \log\left( \frac{392Km}{(\Delta^j(a))^2 \cdot \delta} \right)=\frac{\gamma^j(\delta)\cdot\Delta^i(a)}{(\Delta^j(a))^2}
\end{equation*}
to the $i$-th objective’s regret. Summing over all $j \leq i$ gives the bound:
\begin{equation*}
R^i(t)\leq\sum_{j=1}^i\sum_{a\in \S(j)}\frac{\gamma^j(\delta)\cdot\Delta^i(a)}{(\Delta^j(a))^2}.
\end{equation*}
This completes the proof of Theorem~\ref{thm:1}.
$\hfill\square$

\section{Proof of Theorem~\ref{thm:2}}

With Lemma~\ref{lem:bounded-times} in hand, the proof of Theorem~\ref{thm:2} follows directly. To eliminate any suboptimal arm $a \in \S(i)$, the algorithm requires at most
\begin{equation}
\frac{64}{(\Delta^i(a))^2} \log\left( \frac{392Km}{(\Delta^i(a))^2 \cdot \delta} \right)=\frac{\gamma^i(\delta)}{(\Delta^i(a))^2} 
\end{equation}
pulls. 

To identify the set $\O_*(i)$, the set of arms that are optimal up to objective $i$, the algorithm must eliminate all arms in $\S(j)$ for every $j \leq i$. Therefore, the total number of samples required by LexElim-Out to identify $\O_*(i)$ is bounded by:
\begin{equation*}
T^i(\delta)\leq \sum_{j=1}^i\sum_{a\in \S(j)}\frac{\gamma^j(\delta)}{(\Delta^j(a))^2}.
\end{equation*}
This concludes the proof of Theorem~\ref{thm:2}.
$\hfill\square$

\section{Proof of Theorem~\ref{thm:3}}
To begin with, we prove that the lex-optimal arm $a_*$ is not eliminated during the Steps~6 to 9 in Algorithm~\ref{LexElim-In}.

\begin{lem}\label{lem:optimal-contain}
Suppose $\event$ in Eq.~\eqref{event} holds. In Steps 6 to 9 of Algorithm~\ref{LexElim-In}, if $a_*\in\A_{t}^0$, then 
\begin{equation*}
a_*\in\A_{t}^m \quad \text{and} \quad \Delta^i(a)\leq4(1+\lambda+\cdots+\lambda^{i-1})\cdot c_t(a_t),\quad \forall i\in[m],\ \forall a\in\A_t^m.
\end{equation*}
\end{lem}

\noindent\textbf{Proof:} We prove the lemma via induction on the objective index $i\in[m]$.

\textbf{Base case ($i=1$):} Since $\hat{a}_t^1 = \argmax_{a\in\A_t^0} \hat{\mu}_t^1(a)$ and $a_*\in\A_t^0$, for all $a\in\A_t^1$, we have
\begin{equation}\label{lem2:1}
\Delta^1(a) = \mu^1(a_*) - \mu^1(a) \leq \mu^1(a_*) - \hat{\mu}_t^1(a_*) + \hat{\mu}_t^1(\hat{a}_t^1) - \mu^1(a).
\end{equation}
Under event $\event$, it holds that
\begin{equation*}
\mu^1(a_*) - \hat{\mu}_t^1(a_*) \leq c_t(a_*), \qquad \hat{\mu}_t^1(a) - \mu^1(a) \leq c_t(a), \quad \forall a\in\A_t^1.
\end{equation*}
Plugging these into Eq.~\eqref{lem2:1}, we obtain
\begin{equation*}
\Delta^1(a) \leq c_t(a_*) + \hat{\mu}_t^1(\hat{a}_t^1) - \hat{\mu}_t^1(a) + c_t(a), \quad \forall a\in\A_t^1.
\end{equation*}
By the elimination rule, for all $a\in\A_t^1$,
\begin{equation*}
\hat{\mu}_t^1(\hat{a}_t^1) - \hat{\mu}_t^1(a) \leq 2c_t(a_t).
\end{equation*}
Moreover, since $a_t = \argmax_{a\in\A_t^0} c_t(a)$, it holds that $c_t(a)\leq c_t(a_t)$ and $c_t(a_*) \leq c_t(a_t)$. Hence,
\begin{equation*}
\Delta^1(a) \leq c_t(a_t) + 2c_t(a_t) + c_t(a_t) = 4c_t(a_t), \quad \forall a\in\A_t^1.
\end{equation*}
Finally, since
\begin{equation*}
\hat{\mu}_t^1(\hat{a}_t^1) - \hat{\mu}_t^1(a_*) \leq \mu^1(\hat{a}_t^1) + c_t(\hat{a}_t^1) - \mu^1(a_*) + c_t(a_*) \leq 2c_t(a_t),
\end{equation*}
we conclude that $a_*\in\A_t^1$.

\textbf{Inductive step:} Suppose that for all $j \leq i-1$, it holds that $a_*\in\A_t^j$ and
\begin{equation*}
\Delta^j(a) \leq 4(1+\lambda+\cdots+\lambda^{j-1})\cdot c_t(a_t), \quad \forall a\in\A_t^j.
\end{equation*}
We now prove the statement for $j=i$. Since $\hat{a}_t^i = \argmax_{a\in\A_t^{i-1}} \hat{\mu}_t^i(a)$ and $a_*\in\A_t^{i-1}$, then for all $a\in\A_t^i \subseteq \A_t^{i-1}$,
\begin{equation}\label{lem2:2}
\Delta^i(a) = \mu^i(a_*) - \mu^i(a) \leq \mu^i(a_*) - \hat{\mu}_t^i(a_*) + \hat{\mu}_t^i(\hat{a}_t^i) - \mu^i(a).
\end{equation}
By the event $\event$, we have
\begin{equation}
\mu^i(a_*) - \hat{\mu}_t^i(a_*) \leq c_t(a_*), \qquad \hat{\mu}_t^i(a) - \mu^i(a) \leq c_t(a),\quad \forall a\in\A_t^i.
\end{equation}
Substituting into Eq.~\eqref{lem2:2}, we get
\begin{equation*}
\Delta^i(a) \leq c_t(a_*) + \hat{\mu}_t^i(\hat{a}_t^i) - \hat{\mu}_t^i(a) + c_t(a).
\end{equation*}
From the elimination rule in Algorithm~\ref{LexElim-In}, it follows that
\begin{equation*}
\hat{\mu}_t^i(\hat{a}_t^i) - \hat{\mu}_t^i(a) \leq (2 + 4\lambda + \cdots + 4\lambda^{i-1})\cdot c_t(a_t), \quad \forall a\in\A_t^i.
\end{equation*}
Also, since $a_t = \argmax_{a\in\A_t^0} c_t(a)$, we have $c_t(a), c_t(a_*) \leq c_t(a_t)$, thus
\begin{equation*}
\Delta^i(a) \leq 2c_t(a_t) + (2 + 4\lambda + \cdots + 4\lambda^{i-1})\cdot c_t(a_t) = 4(1+\lambda+\cdots+\lambda^{i-1})\cdot c_t(a_t).
\end{equation*}

Next, we show $a_*\in\A_t^i$. By the same reasoning as above,
\begin{equation*}
\hat{\mu}_t^i(\hat{a}_t^i) - \hat{\mu}_t^i(a_*) \leq \mu^i(\hat{a}_t^i) + c_t(\hat{a}_t^i) - \mu^i(a_*) + c_t(a_*).
\end{equation*}
From the lexicographic trade-off in Eq.~\eqref{lex-identity-MAB} and the inductive assumption,
\begin{equation*}
\mu^i(\hat{a}_t^i) - \mu^i(a_*) \leq \lambda \cdot \max_{j\in[i-1]} \{ \mu^j(a_*) - \mu^j(\hat{a}_t^i) \} \leq \lambda \cdot 4(1+\lambda+\cdots+\lambda^{i-2})\cdot c_t(a_t).
\end{equation*}
Using $c_t(\hat{a}_t^i), c_t(a_*) \leq c_t(a_t)$, it follows that
\begin{equation*}
\hat{\mu}_t^i(\hat{a}_t^i) - \hat{\mu}_t^i(a_*) \leq 4\lambda(1+\lambda+\cdots+\lambda^{i-2})\cdot c_t(a_t) + 2c_t(a_t) = (2 + 4\lambda + \cdots + 4\lambda^{i-1})\cdot c_t(a_t).
\end{equation*}
Thus, $a_*\in\A_t^i$. By induction, this holds for all $i\in[m]$. Therefore, we conclude
\begin{equation*}
a_*\in\A_t^m,\quad \Delta^i(a) \leq 4(1+\lambda+\cdots+\lambda^{i-1})\cdot c_t(a_t),\quad \forall i\in[m],\ \forall a\in\A_t^m.
\end{equation*}
This completes the proof.
$\hfill\square$

Then, we provide an upper bound on the number of times a suboptimal arm can be pulled in the LexElim-In algorithm.

\begin{lem}\label{lem:bounded-times-In}
Suppose the event $\event$ in \eqref{event} holds. In Algorithm~\ref{LexElim-In}, for any arm $a\in[K]$, the number of times it is played is at most
\begin{equation*}
n_t(a) \leq \min_{i\in[m]}\left\{\frac{64(\Lambda^i(\lambda))^2}{(\Delta^i(a))^2\cdot\ID(\Delta^i(a)>0)}\log\left( \frac{392Km}{(\Delta^i(a))^2 \cdot \delta} \right)\right\},
\end{equation*}
where $\Lambda^i(\lambda) = 1 + \lambda + \cdots + \lambda^{i-1}$.
\end{lem}

\noindent\textbf{Proof.} Fix any arm $a\in[K]$. Let $i$ be an index such that $a$ is suboptimal with respect to the $i$-th objective, i.e., $\Delta^i(a) > 0$, and is eliminated based on the reward estimates of objective $i$ in some round $t$.

In Algorithm~\ref{LexElim-In}, an arm $a$ is removed from $\A_t^{i-1}$ according to objective $i$ if
\begin{equation}\label{eq:elim-in}
\hat{\mu}_t^i(\hat{a}_t^i) - \hat{\mu}_t^i(a) > (2 + 4\lambda + \cdots + 4\lambda^{i-1}) \cdot c_t(a_t), 
\end{equation}
where $a_t = \argmax_{a\in \A_t} c_t(a)$ denotes the arm with the largest confidence width.

Since $\hat{a}_t^i = \argmax_{a\in \A_t^{i-1}} \hat{\mu}_t^i(a)$ and by Lemma~\ref{lem:optimal-contain} we know $a_*\in \A_t^{i-1}$, a sufficient condition for \eqref{eq:elim-in} is
\begin{equation}\label{eq:elim-in-suff}
\hat{\mu}_t^i(a_*) - \hat{\mu}_t^i(a) > (2 + 4\lambda + \cdots + 4\lambda^{i-1}) \cdot c_t(a_t). 
\end{equation}
Under the confidence event $\event$, for all $t$ and $a \in [K]$, we have
\begin{equation}\label{eq:conf-bound}
|\hat{\mu}_t^i(a) - \mu^i(a)| \leq c_t(a) \leq c_t(a_t). 
\end{equation}
Using \eqref{eq:conf-bound}, inequality \eqref{eq:elim-in-suff} holds if
\begin{equation}\label{eq:suff-cond}
\mu^i(a_*) - \mu^i(a) > 4(1 + \lambda + \cdots + \lambda^{i-1}) \cdot c_t(a_t). 
\end{equation}
Define $\Lambda^i(\lambda) = 1 + \lambda + \cdots + \lambda^{i-1}$. Then, \eqref{eq:suff-cond} becomes
\begin{equation}\label{eq:delta-ct}
\Delta^i(a) > 4 \Lambda^i(\lambda) \cdot c_t(a_t). 
\end{equation}
Recall the form of the confidence radius:
\begin{equation}\label{eq:conf-form}
c_t(a_t) = \sqrt{ \frac{4}{n_t(a_t)} \log\left( \frac{6Km \cdot n_t(a_t)}{\delta} \right) }. 
\end{equation}
Combining \eqref{eq:delta-ct} and \eqref{eq:conf-form}, we obtain:
\begin{equation}\label{eq:delta-ineq}
\Delta^i(a) > 4\Lambda^i(\lambda) \cdot \sqrt{ \frac{4}{n_t(a_t)} \log\left( \frac{6Km \cdot n_t(a_t)}{\delta} \right) }. 
\end{equation}
By Lemma~\ref{lem:tool2} and Lemma~\ref{lem:tool3}, $c_t(\cdot)$ is decreasing in $n_t$ and $n_t(a_t) \geq n_t(a)-1$, \eqref{eq:delta-ineq} still holds if we replace $n_t(a_t)$ with $n_t(a)-1$:
\begin{equation*}
\Delta^i(a) > 2\Lambda^i(\lambda) \cdot \sqrt{ \frac{4}{n_t(a)-1} \log\left( \frac{6Km \cdot (n_t(a)-1)}{\delta} \right) }. 
\end{equation*}
Squaring both sides yields:
\begin{equation*}
(\Delta^i(a))^2 > \frac{16(\Lambda^i(\lambda))^2}{n_t(a)-1} \log\left( \frac{6Km \cdot (n_t(a)-1)}{\delta} \right). 
\end{equation*}
Rewriting this inequality gives:
\begin{equation}\label{eq:apply-tool1}
n_t(a)-1 > \frac{16(\Lambda^i(\lambda))^2}{(\Delta^i(a))^2} \log\left( \frac{6Km \cdot (n_t(a)-1)}{\delta} \right). 
\end{equation}
To get an explicit bound, apply Lemma~\ref{lem:tool1} with
\begin{equation*}
a = \frac{16(\Lambda^i(\lambda))^2}{(\Delta^i(a))^2}, \quad b = \frac{6Km}{\delta}, \quad x = n_t(a).
\end{equation*}
According to Lemma~\ref{lem:tool1}, inequality \eqref{eq:apply-tool1} holds if
\begin{equation*}
n_t(a) > \frac{16(\Lambda^i(\lambda))^2}{(\Delta^i(a))^2} \log\left( \frac{384Km}{(\Delta^i(a))^2 \cdot \delta} \right) + 1. 
\end{equation*}
Therefore, the number of times arm $a$ is played is at most
\begin{equation*}
n_t(a) \leq \frac{16(\Lambda^i(\lambda))^2}{(\Delta^i(a))^2} \log\left( \frac{392Km}{(\Delta^i(a))^2 \cdot \delta} \right).
\end{equation*}
Since this holds for every $i \in [m]$ with $\Delta^i(a)>0$, we obtain
\begin{equation*}
n_t(a) \leq \min_{i\in[m]}\left\{ \frac{64(\Lambda^i(\lambda))^2}{(\Delta^i(a))^2 \cdot \ID(\Delta^i(a)>0)} \log\left( \frac{392Km}{(\Delta^i(a))^2 \cdot \delta} \right) \right\}. 
\end{equation*}
This concludes the proof of Lemma~\ref{lem:bounded-times-In}.
$\hfill\square$

We now complete the proof of Theorem~\ref{thm:3}. Recall that for each objective $i\in[m]$, he regret arises solely from the suboptimal arms with $\Delta^i(a)>0$.  The contribution of each such arm $a$ to the regret is given by $\Delta^i(a)\cdot n_t(a)$.  Therefore, the cumulative regret for the $i$-th objective can be bounded as follows:
\begin{equation*}
R^i(t)= \sum_{\Delta^i(a)>0}\Delta^i(a)\cdot n_t(a)\leq\sum_{\Delta^i(a)>0}\min_{j\in[m]}\left\{ \frac{64(\Lambda^j(\lambda))^2\Delta^i(a)}{(\Delta^j(a))^2 \cdot \ID(\Delta^j(a)>0)} \log\left( \frac{392Km}{(\Delta^j(a))^2 \cdot \delta} \right) \right\}.
\end{equation*}
Finally, noting that $\gamma^j(\delta)=64\log\left( \frac{392Km}{(\Delta^j(a))^2 \cdot \delta} \right)$, the proof is finished.
$\hfill\square$

\section{Proof of Corollary~\ref{cor:1}}
From Lemma~\ref{lem:optimal-contain}, we know that for any arm $a \in \A_t^m$, the suboptimality gap satisfies
\begin{equation*}
\Delta^i(a) \leq 4(1+\lambda+\cdots+\lambda^{i-1}) \cdot c_t(a_t), \quad \forall i \in [m].
\end{equation*}
Define the scaling factor $\Lambda^i(\lambda) = 1 + \lambda + \cdots + \lambda^{i-1}$. Since $a_t \in \A_{t-1}^m$, it follows that
\begin{equation*}
\Delta^i(a_t) \leq 4\Lambda^i(\lambda) \cdot c_{t-1}(a_{t-1}).
\end{equation*}
By the definition of regret, we have
\begin{equation}\label{eq:regret-lem}
R^i(t) = \sum_{\tau=1}^t \Delta^i(a_\tau) \leq \sum_{\tau=1}^t 4\Lambda^i(\lambda) \cdot c_{\tau-1}(a_{\tau-1}).
\end{equation}
Recall that the confidence radius is defined as
\begin{equation*}
c_t(a_t) = \sqrt{ \frac{4}{n_t(a_t)} \log\left( \frac{6Km \cdot n_t(a_t)}{\delta} \right) }.
\end{equation*}
Substituting the definition of the confidence radius $c_{\tau-1}(a_{\tau-1})$ into Eq.~\eqref{eq:regret-lem}, we obtain:
\begin{equation*}
R^i(t)\leq \sum_{\tau=1}^t 4\Lambda^i(\lambda) \sqrt{ \frac{4}{n_{\tau-1}(a_{\tau-1})} \log\left( \frac{6Kmt}{\delta} \right) }.
\end{equation*}
We regroup the terms by arm $a \in [K]$ and the number of times each arm has been pulled up to round $t$:
\begin{equation}\label{eq:regret-lem-2}
R^i(t)\leq \sum_{a \in [K]} \sum_{n=1}^{n_{t-1}(a)} 4\Lambda^i(\lambda) \sqrt{ \frac{4}{n} \log\left( \frac{6Kmt}{\delta} \right) }.
\end{equation}

Using the standard inequality
\begin{equation*}
\sum_{n=1}^{N} \frac{1}{\sqrt{n}} \leq 2\sqrt{N},
\end{equation*}
we upper-bound the inner sum of Eq.~\eqref{eq:regret-lem-2} as
\begin{equation*}
R^i(t) \leq \sum_{a \in [K]} 8\Lambda^i(\lambda) \cdot \sqrt{4 \cdot n_{t-1}(a) \cdot \log\left( \frac{6Kmt}{\delta} \right) }.
\end{equation*}
Simplifying constants, we arrive at,
\begin{equation*}
R^i(t) \leq \sum_{a \in [K]} 16\Lambda^i(\lambda) \sqrt{ n_{t-1}(a) \cdot \log\left( \frac{6Kmt}{\delta} \right) }.
\end{equation*}

Finally, applying Jensen's inequality (or concavity of the square root), we bound the total sum
\begin{equation*}
\sum_{a \in [K]} \sqrt{n_{t-1}(a)} \leq \sqrt{K \cdot \sum_{a} n_{t-1}(a)} \leq \sqrt{K t}.
\end{equation*}
Therefore, the regret is bounded as
\begin{equation*}
R^i(t) \leq 16\Lambda^i(\lambda) \cdot \sqrt{Kt \cdot \log\left( \frac{6Kmt}{\delta} \right) } = \widetilde{O}(\Lambda^i(\lambda) \cdot \sqrt{K t}).
\end{equation*}
The proof of Corollary~\ref{cor:1} is finished.
$\hfill\square$

\section{Proof of Theorem~\ref{thm:4}}

With Lemma~\ref{lem:bounded-times-In} in hand, the proof of Theorem~\ref{thm:4} follows directly. To eliminate any suboptimal arm $a$ that $\Delta^i(a)>0$, the algorithm requires at most
\begin{equation*}
\min_{j\in[m]}\left\{ \frac{64(\Lambda^j(\lambda))^2}{(\Delta^j(a))^2 \cdot \ID(\Delta^j(a)>0)} \log\left( \frac{392Km}{(\Delta^j(a))^2 \cdot \delta} \right) \right\}
\end{equation*}
pulls. 

To identify the set $\widetilde{\O}_*(i) = \{a \in [K] \mid \Delta^i(a)\leq 0\}$, the set of arms that are optimal up to objective $i$, the algorithm must eliminate all arms $\Delta^i(a)>0$. Therefore, the total number of samples required by LexElim-In to identify $\widetilde{\O}_*(i)$ is bounded by:
\begin{equation*}
\widetilde{T}^i(\delta)\leq \sum_{\Delta^i(a)>0}\min_{j\in[m]}\left\{\frac{(\Lambda^j(\lambda))^2\cdot \gamma^j(\delta)}{(\Delta^j(a))^2\cdot\ID(\Delta^j(a)>0)}\right\}, \gamma^j(\delta)=64\log\left( \frac{392Km}{(\Delta^j(a))^2 \cdot \delta} \right).
\end{equation*}
This concludes the proof.
$\hfill\square$

\section{Proof of Technical Lemmas}

\setcounter{lem}{0} 

\begin{lem}
With probability at least $1-\delta$, for any $t\geq1$, 
\begin{equation*}
\left|\hat{\mu}_t^i(a)-\mu^i(a)\right|\leq c_t(a), i\in [m], a \in [K].
\end{equation*}
\end{lem}

\noindent\textbf{Proof.} If $n_t(a) = 0$, then by definition $c_t(a) = +\infty$, the inequality holds trivially. We therefore consider the case $n_t(a) \geq 1$.

Fix any objective $i \in [m]$, according to Lemma~6 of \citet{Abbasi:2011}, we have that with probability at least $1 - \delta$, for any $t \geq 1$ and any arm $a \in [K]$, the empirical mean satisfies:
\begin{equation*}
\left|\frac{1}{n_t(a)}\sum_{\tau=1}^{t-1} r_\tau^i(a_\tau)\ID(a_\tau=a)-\mu^i(a)\right|\leq \sqrt{\left(1+2\log\left(\frac{K\sqrt{1+n_t(a)}}{\delta}\right)\right)\frac{1+n_t(a)}{n^2_t(a)}}.
\end{equation*}
Noting that $\hat{\mu}_t^i(a) = \frac{1}{n_t(a)} \sum_{\tau=1}^{t-1} r_\tau^i(a_\tau) \cdot \mathbb{I}(a_\tau = a)$, the above bound directly applies to $\left| \hat{\mu}_t^i(a) - \mu^i(a) \right|$.

Applying a union bound over all $m$ objectives, and replacing $\delta$ with $\delta /m$, we get that with probability at least $1-\delta$, for all $i \in [m]$, $a \in [K]$, and $t \geq 1$,
\begin{equation}\label{eq:union_bound}
\left|\hat{\mu}_t^i(a)-\mu^i(a)\right|\leq \sqrt{\left(1+2\log\left(\frac{Km\sqrt{1+n_t(a)}}{\delta}\right)\right)\frac{1+n_t(a)}{n^2_t(a)}}.
\end{equation}

Using the inequality $\log(Km\cdot \sqrt{e}\cdot \sqrt{1 + n_t(a)}/ \delta) \leq \log(6Km \cdot n_t(a)/\delta)$ for $n_t(a) \geq 1$, we can further relax the bound in Eq.~\eqref{eq:union_bound} to:
\begin{equation*}
\left| \hat{\mu}_t^i(a) - \mu^i(a) \right|
\leq \sqrt{ \frac{4}{n_t(a)} \log\left( \frac{6Km \cdot n_t(a)}{\delta} \right) } =: c_t(a).
\end{equation*}
This completes the proof.
$\hfill\square$

\begin{lem}
Let $a > 0$, $b > 0$ and $ab>e$. If $x > a \log(ab)$, then $x > a \log(bx)$.
\end{lem}
\noindent\textbf{Proof.} Define the function $f(x) = x - a \log x$. We aim to find a value $x_0$ such that $f(x_0) > a \log b$, which implies
\begin{equation*}
x_0 - a \log x_0 > a \log b \quad \Leftrightarrow \quad x_0 > a \log(bx_0).
\end{equation*}

First, observe that $f(x)$ is differentiable and its derivative is given by
\begin{equation*}
f'(x) = 1 - \frac{a}{x}.
\end{equation*}
Thus, $f(x)$ is strictly increasing for all $x > a$.

Now, let us consider $x_0 = a \log(ab)$. Note that $\log(ab) = \log a + \log b$, and so $x_0 = a (\log a + \log b)$. We compute
\begin{equation*}
a \log(bx_0) = a \log\left(ba \log(ab)\right) = a(\log a + \log b + \log\log(ab)).
\end{equation*}
Since $\log\log(ab) < \log(ab)$ for all $ab > e$, it follows that
\begin{equation*}
x_0 = a \log(ab) > a \log(bx_0).
\end{equation*}
Hence, $x_0$ satisfies the inequality, and due to the monotonicity of $f(x)$ for $x > a$, any $x > x_0$ also satisfies
\begin{equation*}
x > a \log(bx).
\end{equation*}
The proof is finished.
$\hfill\square$

\begin{lem}
Let $f(n) = 4\sqrt{ \frac{4}{n} \log\left( \frac{6Km \cdot n}{\delta} \right) }$ for $n > 0$. Then, $f(n)$ is strictly decreasing for all
\begin{equation*}
n > \frac{e\delta}{6Km}.
\end{equation*}
In particular, since $\frac{e\delta}{6Km} \ll 1$ in typical applications, the function $f(n)$ is strictly decreasing for all $n \geq 1$.
\end{lem}
\noindent\textbf{Proof.} Let $C = \frac{6Km}{\delta}$, so that the function becomes:
\begin{equation*}
f(n) = 8\sqrt{ \frac{ \log(Cn) }{ n } }.
\end{equation*}
Define the inner function $h(n) = \frac{\log(Cn)}{n}$, so that $f(n) = 8\sqrt{h(n)}$. It suffices to show that $h(n)$ is strictly decreasing. Taking the derivative:
\begin{equation*}
h'(n) = \frac{1 - \log(Cn)}{n^2}.
\end{equation*}
Hence, $h'(n) < 0$ if and only if $\log(Cn) > 1$, which is equivalent to $Cn > e$. Therefore, $f(n)$ is strictly decreasing for all $n > \frac{e}{C} = \frac{e\delta}{6Km}$, as claimed.
$\hfill\square$

\end{document}